\documentclass{article}
\usepackage{spconf,amsmath,graphicx}
\usepackage{booktabs}
\usepackage{graphicx}
\usepackage[T1]{fontenc} 
\usepackage{hyperref}

\title{Multimodal Information Interaction for Medical Image Segmentation.}
\name{Xinxin Fan$^{1,2,\dagger}$ , Lin Liu$^{1,2,\dagger}$,Haoran Zhang$^{1}$,\thanks{*Corresponding author.}\thanks{$^{\dagger}$First Author and Second Author contribute equally to this work.}}
\address{$^1$Shenzhen Institute of Advanced Technology, Chinese Academy of Sciences, Shenzhen, China  \\
	$^2$ University of Chinese Academy of Sciences, Beijing, China \\
	xx.fan@siat.ac.cn
}

\begin{document}
%
\maketitle
\begin{abstract}
The use of multimodal data in assisted diagnosis and segmentation has emerged as a prominent area of interest in current research. However, one of the primary challenges is how to effectively fuse multimodal features. Most of the current approaches focus on the integration of multimodal features while ignoring the correlation and consistency between different modal features, leading to the inclusion of potentially irrelevant information. 
To address this issue, we introduce an innovative Multimodal Information Cross Transformer (MicFormer), which employs a dual-stream architecture to simultaneously extract features from each modality. Leveraging the Cross Transformer, it queries features from one modality and retrieves corresponding responses from another, facilitating effective communication between bimodal features. Additionally, we incorporate a deformable Transformer architecture to expand the search space. We conducted experiments on the MM-WHS dataset, and in the CT-MRI multimodal image segmentation task, we successfully improved the whole-heart segmentation DICE score to 85.57 and MIoU to 75.51. Compared to other multimodal segmentation techniques, our method outperforms by margins of 2.83 and 4.23, respectively. This demonstrates the efficacy of MicFormer in integrating relevant information between different modalities in multimodal tasks. These findings hold significant implications for multimodal image tasks, and we believe that MicFormer possesses extensive potential for broader applications across various domains. Access to our method is available at \href{https://github.com/fxxJuses/MICFormer}{https://github.com/fxxJuses/MICFormer}.
\end{abstract}
\section{Introduction}
\label{sec:intro}
\begin{figure}[htbp]
	\centering
	\includegraphics[width=80mm]{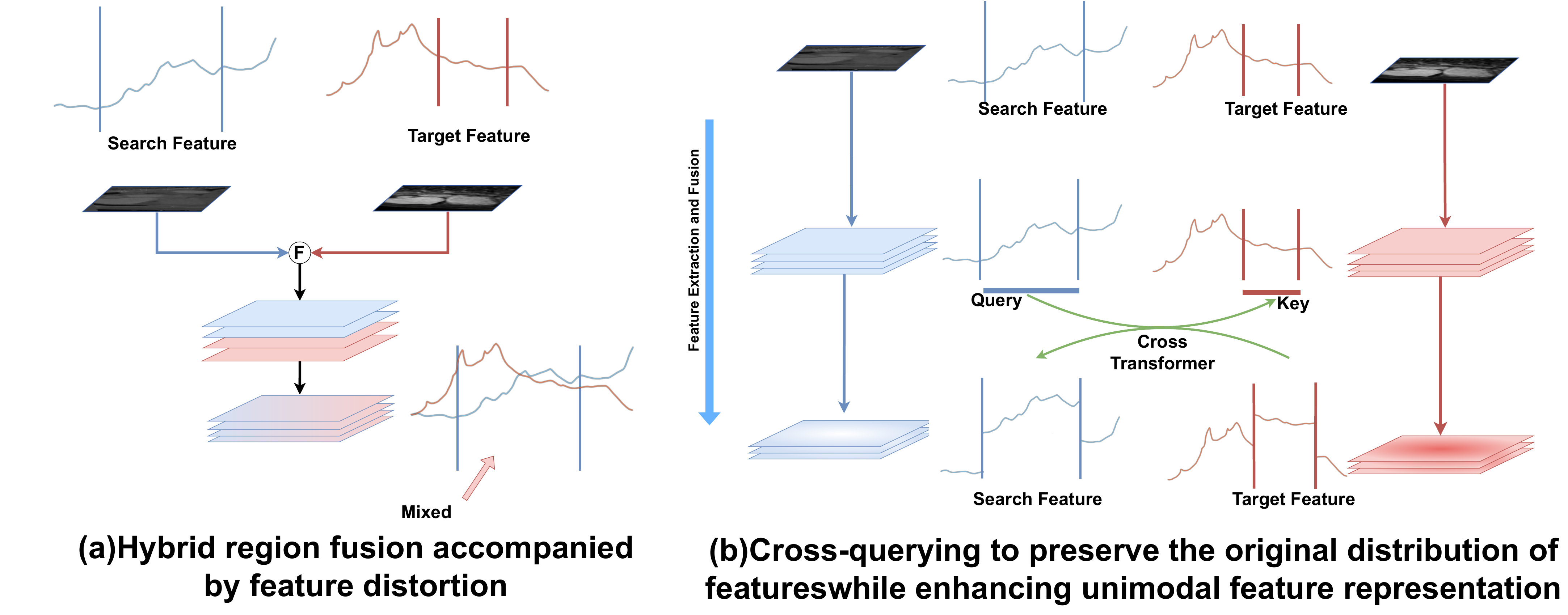}
	\caption{Limitations of the Current Method: (a) The prioritization of image fusion using unimodal network form for multimodal image segmentation can result in inaccurate feature representation in the target feature region. (b) Modal feature fusion is restricted to a two-stream cross-attention fusion network. Query and Key matching is used to enhance the unimodal feature representation without incorporating additional information. Additionally, technical term abbreviations will be explained when first used.}
	\label{Fig1}
\end{figure} 
Multimodal medical image segmentation capitalizes on the integration of information from diverse imaging modalities to achieve superior segmentation precision\cite{Liu2023AMC,Yang2022TowardUM,Dolz2018HyperDenseNetAH}. For example, in the context of whole-heart segmentation, augmenting CT-based segmentation with MRI data can significantly improve the accuracy of the CT segmentation\cite{Cui2023AnIC,Xu2018CFUNCF}. This improvement can be attributed to the enhanced contrast and superior soft tissue imaging characteristics inherent in MRI. State-of-the-art multimodal segmentation networks\cite{Lin2022CKDTransBTSCK,Yang2023FlexibleFN,10039141} are designed to proficiently assimilate information from these disparate modalities. They accomplish this by extracting salient features from two modalities and subsequently merging the relevant features to produce segmentations of exceptional precision.

While current deep learning networks exhibit commendable performance on multimodal segmentation tasks, a majority of the techniques still predominantly employ a single-stream approach for feature alignment and fusion. As depicted in Figure \ref{Fig1}, several methods (including TransUnet\cite{Chen2021TransUNetTM}, Swin-Unet\cite{Cao2021SwinUnetUP}, and MedNeXt\cite{Roy2023MedNeXtTS}) treat the multimodal image pairs as a unified image input. Consequently, these methods place complete reliance on the network's internal mechanisms, such as feature extraction, to execute feature fusion. This methodology fails to establish an explicit correspondence between different modalities, inadvertently introducing redundant information into the primary features. Such perturbations pose challenges in refining these features in subsequent feature extraction stages.

Transformer and its derivatives\cite{Chen2021TransUNetTM,Cao2021SwinUnetUP,peiris2022robust,Hatamizadeh2022SwinUS,Zhou2021nnFormerIT} are more concerned with self-attention and lack of correlation attention between image pairs. Furthermore, the Swin Transformer\cite{Liu2021SwinTH} design is constrained by its necessity to search within a predetermined corresponding window. Given that multimodal features often exhibit a greater degree of positional variability, Swin Transformer architectures may not be well-suited for extensive cross-window searches. Thus, the development of an attention mechanism tailored for the matching of multimodal image features could substantially enhance the efficiency of multimodal segmentation tasks.

In this paper, we propose a novel multimodal transformer, MicFormer. For handling feature fusion and matching in multimodal tasks. By leveraging a dual-stream model, MicFormer adeptly abstracts from both modalities. Central to its efficacy is the Deformable Cross Attention module, which advances the querying and communication between the two modal features. The architecture that emerges ensures a competent fusion of multi-level, multimodal semantic features, providing significant advances in multimodal segmentation accuracy. In short, the pivotal contributions of our work are:1) We propose a novel two-stream multimodal feature fusion transformer backbone network. The multimodal features are continuously matched by using a deformable cross-attention fusion module.2) We propose a novel deformable cross-attention module for the automatic adaptation of the search space.

\section{Method}
\label{sec:format}

\begin{figure*}[htbp]
	\centering
	\includegraphics[width=160mm]{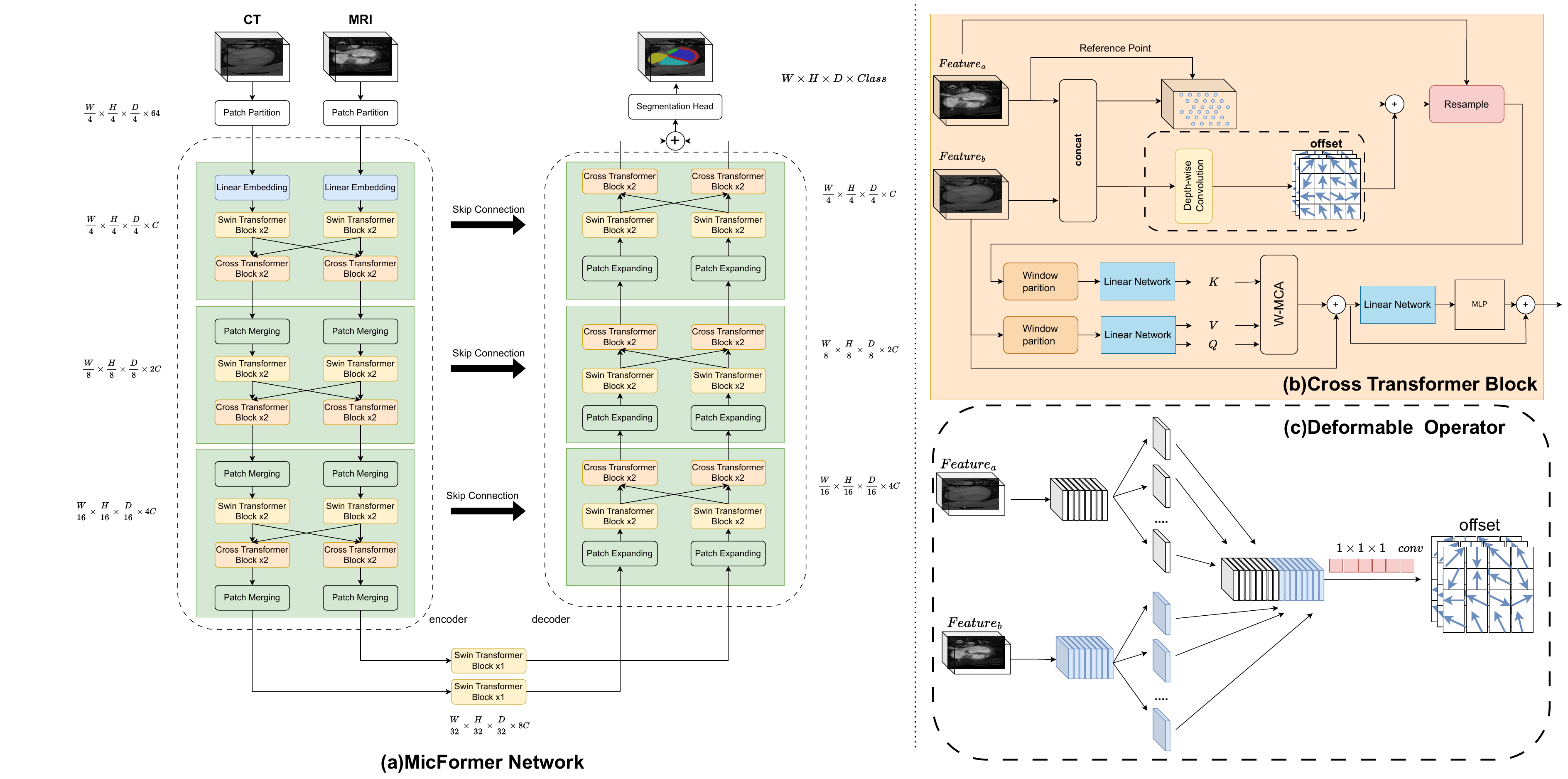}
	\caption{Our MicFormer architecture, which consists of (a) U-shaped parallel feature network.(b) Cross Transformer Block.(c)Deformable Operator. We utilize deep separable convolution to split $Feature_a$ and $Feature_b$ in the channel direction and compute the positional differences at the corresponding positions.}
	\label{Fig2}
\end{figure*}

As depicted in Figure \ref{Fig2}, the architecture of MicFormer is rooted in a Swin-Unet deep learning segmentation network, further enhanced by the incorporation of a parallel two-stream architecture. MicFormer itself comprises a Transformer architecture alongside a parallel sub-network featuring U-shaped feature extraction, thereby facilitating feature fusion and enabling a continuous module exchange between modalities through the utilization of a deformable cross-attention mechanism. Within this framework, the Cross Transformer module employs deformable sampling to compute the structural relationship between both modalities, thus reshaping one modality's structural information to align with the corresponding structures of both modalities within the same local window of the Swin Transformer. This alignment leads to an improved calculation of similarity. The Cross Transformer module serves to uphold the original feature distribution of the modalities and accentuates the pertinent segmentation features while mitigating the influence of extraneous features by integrating both modalities. Specifically, it augments the segmentation-relevant features while diminishing the significance of non-segmentation-related ones through their reciprocal interaction.

\subsection{U-shaped parallel feature network}
We introduce a U-shaped network designed for concurrent feature extraction, thereby facilitating the seamless integration of information from two distinct modalities. As illustrated in Figure \ref{Fig2}, the two parallel networks adhere to the architectural structure reminiscent of U-net, encompassing both encoding and decoding components. Notably, conventional convolution operations are replaced with a composite block comprising Swin Transformer and Cross Transformer, which are alternately applied. The Swin Transformer's intrinsic capacity to capture both global and local information is harnessed for feature extraction from each modality, followed by cross-modal feature querying in the Cross Transformer. This query process serves to enhance the feature representation within a single modality. MicFormer employs a parallel communication network to continually leverage information from the complementary modality, thereby augmenting the feature expression capabilities within its own modality. Consequently, the resulting output features exhibit multimodal enhancement.

\subsection{Cross Transformer block}
As depicted in Figure \ref{Fig2}(b), the features denoted as $Feature_a$ and $Feature_b$, emanating from the parallel sub-network, engage in a bidirectional querying process within the Cross Transformer block. This process entails the exchange of sequential inputs, thereby mutually enhancing their respective features, which are subsequently reintegrated into the original input pipeline to undergo further rounds of deep feature extraction and interaction.   Given the inherent limitations of the receptive field, attaining a sufficient number of matches within a single instance proves unfeasible. Consequently, the Cross Transformer block adopts a sequential execution strategy involving two distinct communication phases. This approach ensures that our MicFormer not only facilitates substantial unimodal feature extraction but also supports iterative feature refinement through numerous cross-queries.

\subsection{Cross Attention}
The primary objective of the cross-attention mechanism is to establish correlations between the two modalities by harnessing the power of the attention mechanism. This is realized by computing the correlation between the features within $Feature_a$ and those within $Feature_b$. Please refer to Figure \ref{Fig2} for a visual representation of this process. Specifically, we leverage the encoded vectors of $Feature_b$ as both Query and Value, while the encoded vectors of $Feature_a$ serve as Key. The Query operation facilitates the mapping of correlations between individual points in $Feature_b$ and the corresponding points in $Feature_a$ through correlation calculations with the Key.
Subsequently, we apply a Softmax calculation to the correlation map, denoted as $Attn$, which assigns higher weights to features that may be challenging to capture within $Feature_b$ but are more readily discernible in $Feature_a$. As an example, in the context of CT features, the delineation of tissue edges can often be indistinct and challenging, whereas MRI features tend to offer clearer distinctions in tissue edges. Consequently, when calculating the similarity between CT features and MRI tissue edges, higher similarity values are allocated to the edge regions of CT features. This strategic emphasis enables the model to devote more attention to the original, less distinct tissue edges within CT scans.

Ultimately, the Key is subjected to a matrix multiplication with the Softmax correlation mapping $Attn$, ensuring that features within $Feature_a$ that are more challenging to discern within $Feature_b$ receive heightened focus and attention from the model. Our computation of Cross Attention is detailed as follows: 

\begin{equation}
	W-MCA = Softmax(\frac{Q_bK_a^T}{\sqrt{d}})V_b
\end{equation}

Where $Q_b$, $K_a$ and $V_b$ are Query, Key, Value matrices, respectively, $Q_b$ and $V_b$ come from the encoding of $Feature_b$ and $k_a$ comes from the encoding of $Feature_a$.where $d$ represents the feature dimension of the Key.

\subsection{Deformable Operator}
Given that the Swin Transformer relies on a fixed window shape and requires a broader and scattered exploration for cross-modal feature querying, we choose to avoid computationally intensive global searches similar to those in the Vision Transformer model\cite{Dosovitskiy2020AnII}. Instead, we introduce a deformable operation, as shown in Figure \ref{Fig2}, which uses deep convolution to establish relationships between voxel positions in the local neighborhood.  This operation calculates a unified deformation trend called $offset$, which dynamically adjusts to accommodate all features. The $offset$ is a tensor with three channels, representing the displacement of a point on the feature map in the $x$, $y$, and $z$ directions. Importantly, the shape of the $offset$ tensor match those of the feature map.

In the final step, we resample $Feature_a$ to obtain the deformed feature map, adjusting its position accordingly.  This process is similar to applying the same deformable operation to the original queried window. By introducing the Deformable Operator, we effectively transform the Swin Transformer with a fixed receptive field into an equivalent model with a deformable receptive field.

\section{Experiments and Results}
\label{sec:pagestyle}

\subsection{Dataset and Evaluation Metrics}
The MMWHS dataset\cite{Zhuang2016MultivariateMM} consists of 20 cardiac MRI samples with expertannotations for seven structures: left and right ventricles, left and right atrium, pulmonary artery, myocardium, and aorta. pulmonary artery, myocardium, and aorta. In this experiment, we used the SyN algorithm\cite{Avants2020AdvancedNT} CT-MRI image pairs for alignment and cropped the corresponding ROI regions. We randomly cut the entire dataset, with 16 pairs in the training set and 4 cases in the test set.Moreover, the Dice similarity coefficient (Dice), mean intersection over union (MIoU), and 95\% Hausdorff distance (HD95) are employed to evaluate the model performance.

\subsection{Implementation Details}
We use Pytorch to train our MicFormer on NVIDIA RTX 2080 GPU. During the training process, we optimize MicFormer using Adam with a learning rate of 0.0001. The batch size is set as 1, and the maximum epoch is 1000.

\subsection{Compare with the State-of-the-art Methods}

\begin{table}[h]
	\centering
	\renewcommand{\arraystretch}{1.4} 
	\setlength{\tabcolsep}{5pt} 
	\caption{Comparison with different network architectures}
	\resizebox{\linewidth}{!}{
		\begin{tabular}{lcccc}
			\toprule
			Method & Single Modal-/Multi Modal- & Dice $\uparrow$ & MIoU $\uparrow$ & HD95 $\downarrow$ \\
			\midrule
			nnFormer\cite{Zhou2021nnFormerIT}  & Single Modal & 73.67 & 59.73 & 29.12 \\
			\midrule
			VT-Unet \cite{peiris2022robust} & Multi Modal & 77.67 & 64.36 & 14.18   \\
			Swin-Unet \cite{Cao2021SwinUnetUP} & - & 80.22 & 67.59 & 26.92 \\
			SwinUneter \cite{Hatamizadeh2022SwinUS} & - & 81.49 & 69.55 & 14.64  \\
			nnFormer \cite{Zhou2021nnFormerIT} & - & 82.33 & 70.57 & 19.04  \\
			MedNeXt \cite{Roy2023MedNeXtTS} & - & 82.74 & 71.48 & \textbf{10.10}  \\
			MicFormer(ours) & - & \textbf{85.57} & \textbf{75.71} & 10.35  \\
			\bottomrule
		\end{tabular}
	}
	\label{Tabel1}
\end{table}

\begin{figure}[htbp]
	\centering
	\includegraphics[width=70mm]{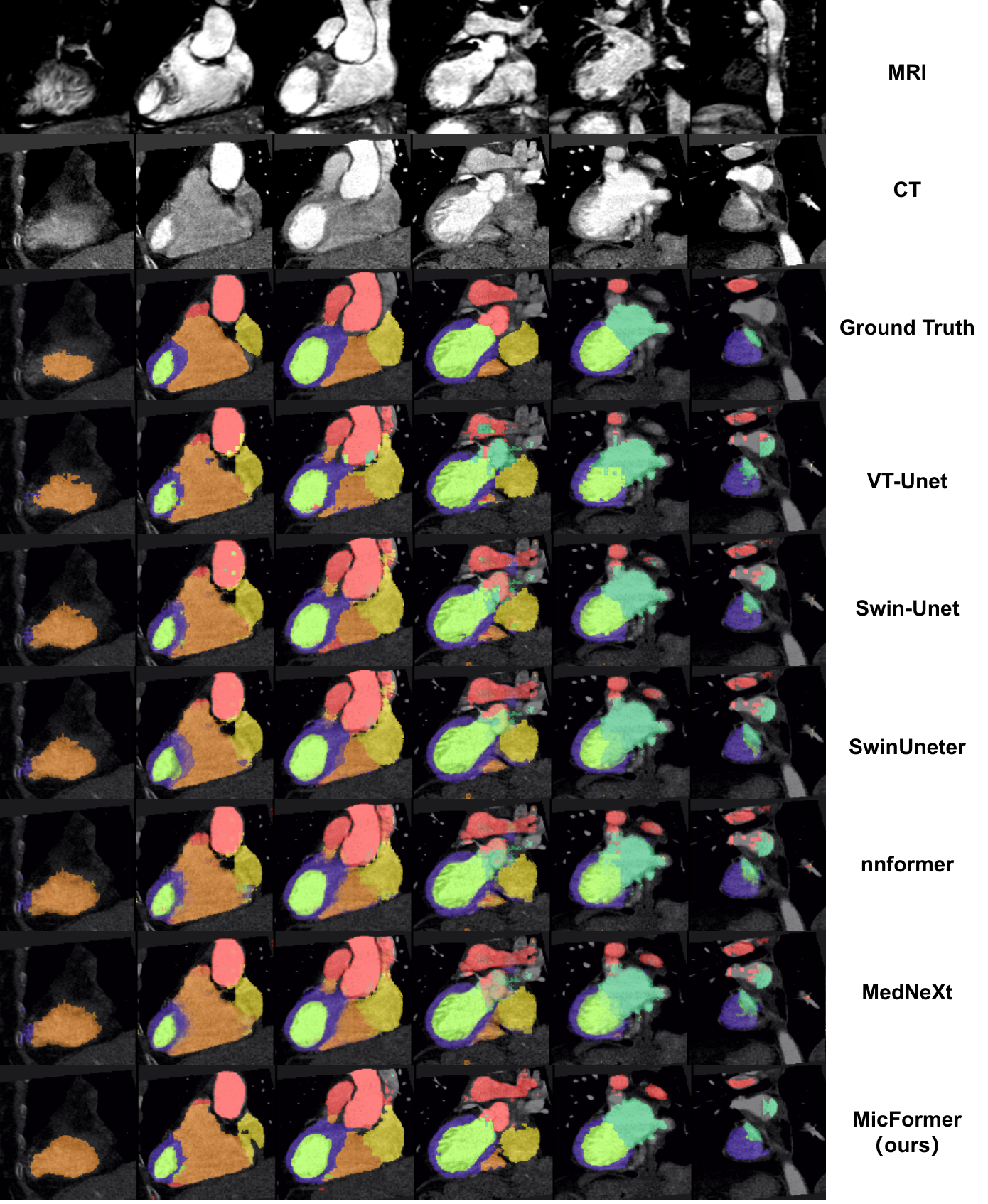}
	\caption{Qualitative Results Analysis: The first row shows MRI image slices, followed by CT image slices in the second row. The third row displays CT's Ground Truth label. The next six rows exhibit the qualitative segmentation results of VT-UNet, Swin-Unet, SwinUneter, nnformer, MedNeXt, and MicFormer.}
	\label{Fig3}
\end{figure} 

We conducted a comprehensive comparison between MicFormer and five state-of-the-art multimodal segmentation algorithms, namely VT-Unet\cite{peiris2022robust}, Swin-Unet\cite{Cao2021SwinUnetUP}, SwinUneter \cite{Hatamizadeh2022SwinUS}, nnFormer\cite{Zhou2021nnFormerIT}, and MedNeXt\cite{Roy2023MedNeXtTS}, with the specific results presented in Table \ref{Tabel1} and Figure \ref{Fig3}. According to the data presented in Table \ref{Tabel1}, MicFormer surpasses all other algorithms in terms of both the Dice coefficient and MIoU. Notably, our results exhibit an improvement of 2.83 and 4.23 points over MedNeXt, respectively, underscoring the superior performance of our segmentation model. However, it is worth mentioning that our method slightly lags behind MedNeXt in terms of the HD95 metric. This discrepancy can be attributed to MedNeXt's utilization of the ConvNeXt architecture\cite{Liu2022ACF}, which exhibits a more robust inductive bias for small datasets and greater sensitivity to boundary information compared to the Transformer architecture. Nevertheless, our method outperforms all other algorithms, with the exception of MedNeXt, on the HD95 metric. This study substantiates that MicFormer excels in capturing segmented surface information across both modalities when compared to the same Transformer architecture. Additionally, our experiments involving Single Modal and Multi Modal scenarios, utilizing nnFormer on the dataset, provide further evidence that the incorporation of MRI data into the CT segmentation process significantly enhances the accuracy of CT segmentation.

As demonstrated in Figure \ref{Fig3}, the performance of various methods on the test set is visualized. Our MicFormer exhibits superior discrimination at tissue edges and tissue-tissue junctions. In the connected domain of the segmentation graph, our method does not display discrete points like VT-Unet, Swin-Unet, SwinUneter and nnFormer.

\section{Conclusion}
\label{sec:majhead}

In this study, we introduce a novel dual-stream cross-network, MicFormer, composed of three main components: a U-shaped parallel feature network, Swin Transformer, and Cross Transformer. The U-shaped parallel feature network prevents uncontrolled fusion by isolating the two modalities. The Swin Transformer captures both global and local information for unimodal feature extraction, while the Cross Transformer employs a deformable cross-attention mechanism to regulate the interaction between the two modalities and enhance the unimodal feature representation. Finally, on the MM-WHS dataset, our approach significantly outperformed other algorithms.



\bibliographystyle{IEEEbib}
\bibliography{strings,refs}

\end{document}